\def\year{2018}\relax
\documentclass[letterpaper]{article} %DO NOT CHANGE THIS
\usepackage{aaai18}  %Required
\usepackage{times}  %Required
\usepackage{helvet}  %Required
\usepackage{courier}  %Required
\usepackage{url}  %Required
\usepackage{graphicx}  %Required
\usepackage{enumerate}
\usepackage{subcaption}
\usepackage{mathtools}

\newcommand{\com}{\text{,}}
\newcommand{\fs}{\text{.}}
\newcommand{\ie}{\mbox{\emph{i.e.\ }}}
\newcommand{\eg}{\mbox{\emph{e.g.\ }}}
\begin{document}
\title{Dual-reference Face Retrieval}
\author{
BingZhang Hu,\textsuperscript{1}
Feng Zheng\textsuperscript{2}
and Ling Shao \textsuperscript{3,1}\\
\textsuperscript{1}{School of Computing Sciences, University of East Anglia, Norwich, UK}\\
\textsuperscript{2}{Department of Electrical and Computer Engineering, University of Pittsburgh, Pittsburgh, USA}\\
\textsuperscript{3}{JD Artificial Intelligence Research (JDAIR), Beijing, China}\\
bingzhang.hu@uea.ac.uk,
feng.zheng@pitt.edu,
ling.shao@ieee.org
}
\maketitle
\begin{abstract}
Face retrieval has received much attention over the past few decades, and many efforts have been made in retrieving face images against pose, illumination, and expression variations. However, the conventional works fail to meet the requirements of a potential and novel task --- retrieving a person's face image at a specific age, especially when the specific `age' is not given as a numeral, \ie `retrieving someone's image at the similar age period shown by another person's image'. To tackle this problem, we propose a dual reference face retrieval framework in this paper, where the system takes two inputs: an identity reference image which indicates the target identity and an age reference image which reflects the target age. In our framework, the raw images are first projected on a joint manifold, which preserves both the age and identity locality. Then two similarity metrics of age and identity are exploited and optimized by utilizing our proposed quartet-based model. The experiments show promising results, outperforming hierarchical methods.
\end{abstract}

%------------------------- Introduction -------------------%
\section{Introduction}
Over the past few decades, face retrieval has received great interest in the research community for its potential applications such as finding missing persons \cite{jain2012face} and matching criminals with CCTV footage for law enforcement \cite{tang2002face}. Apart from a pinch of face retrieval works \cite{bhattacharjee2011construction} that are text-based, most existing frameworks \cite{luo2016tree,lin2017discriminative} are based on the content, in which a target person's image is required as the query input, and the system retrieves all the images belong to the target person in the database. Though these works in some kind improved the benchmark in the past, they fail to catch up with the pace of the new demands of the face retrieval in the age of big data. For example, rather than retrieving all the query identity's images indiscriminately, we may prefer picking out the specific ones with some certain attribute, \eg age. The huge volume of online images makes this kind of fine-grained face retrieval both feasible and indispensable. It is feasible as such large-scale dataset can contain many images taken from someone's different age periods, thereby it is necessary to select them out in some potential applications. 

\begin{figure*}[htp]
	\begin{subfigure}[t]{0.49\textwidth}
			\centering
			\includegraphics[width=\textwidth]{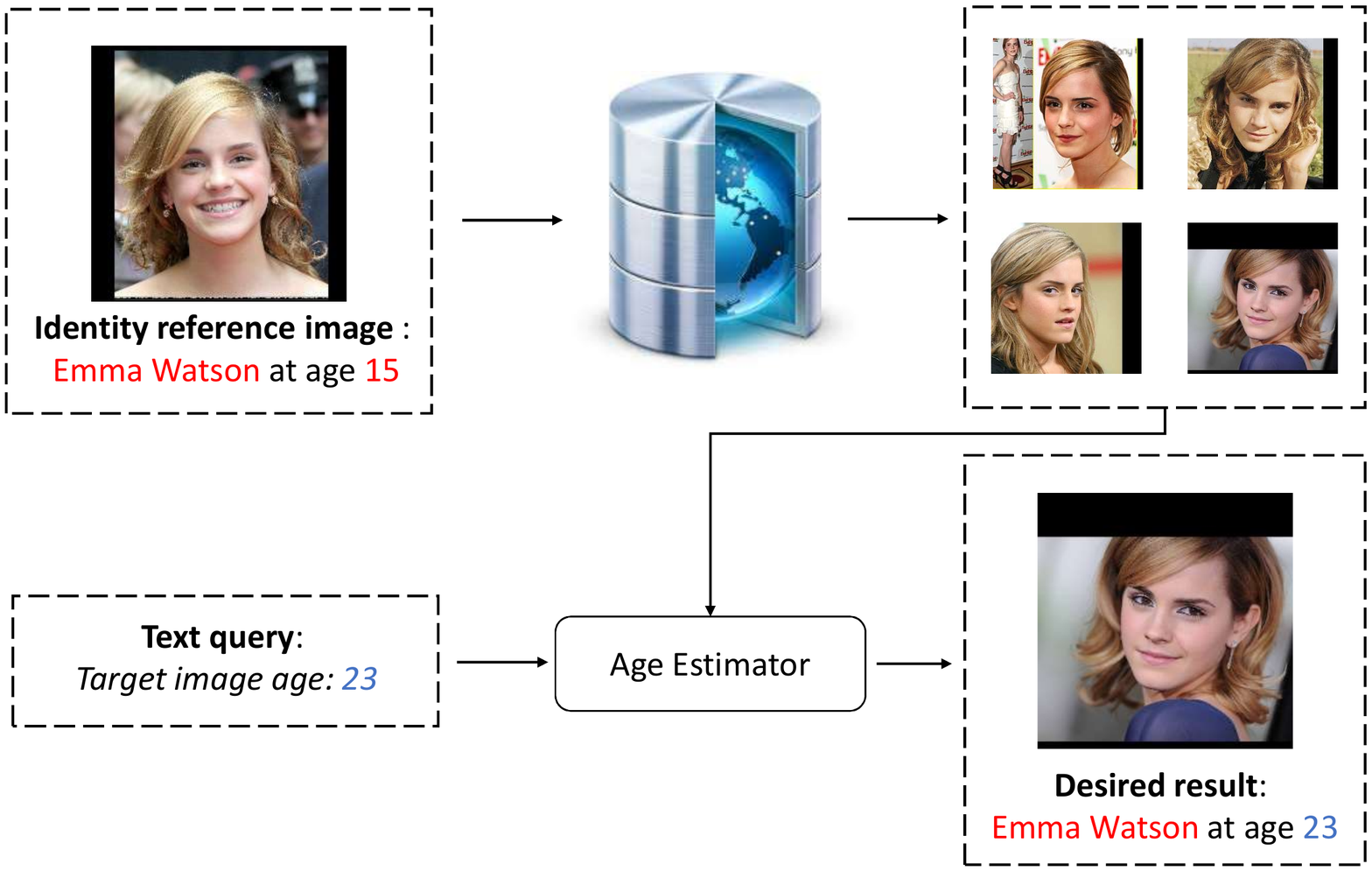}\\
			\caption{Conventional face retrieval framework: Firstly all the images similar to \textit{Emma Watson} are selected, then an external age estimator is employed to select images at the desired age, which is given as a numeral.}
	\end{subfigure}
	\quad
	\begin{subfigure}[t]{0.49\textwidth}
		\centering
		\includegraphics[width=\textwidth]{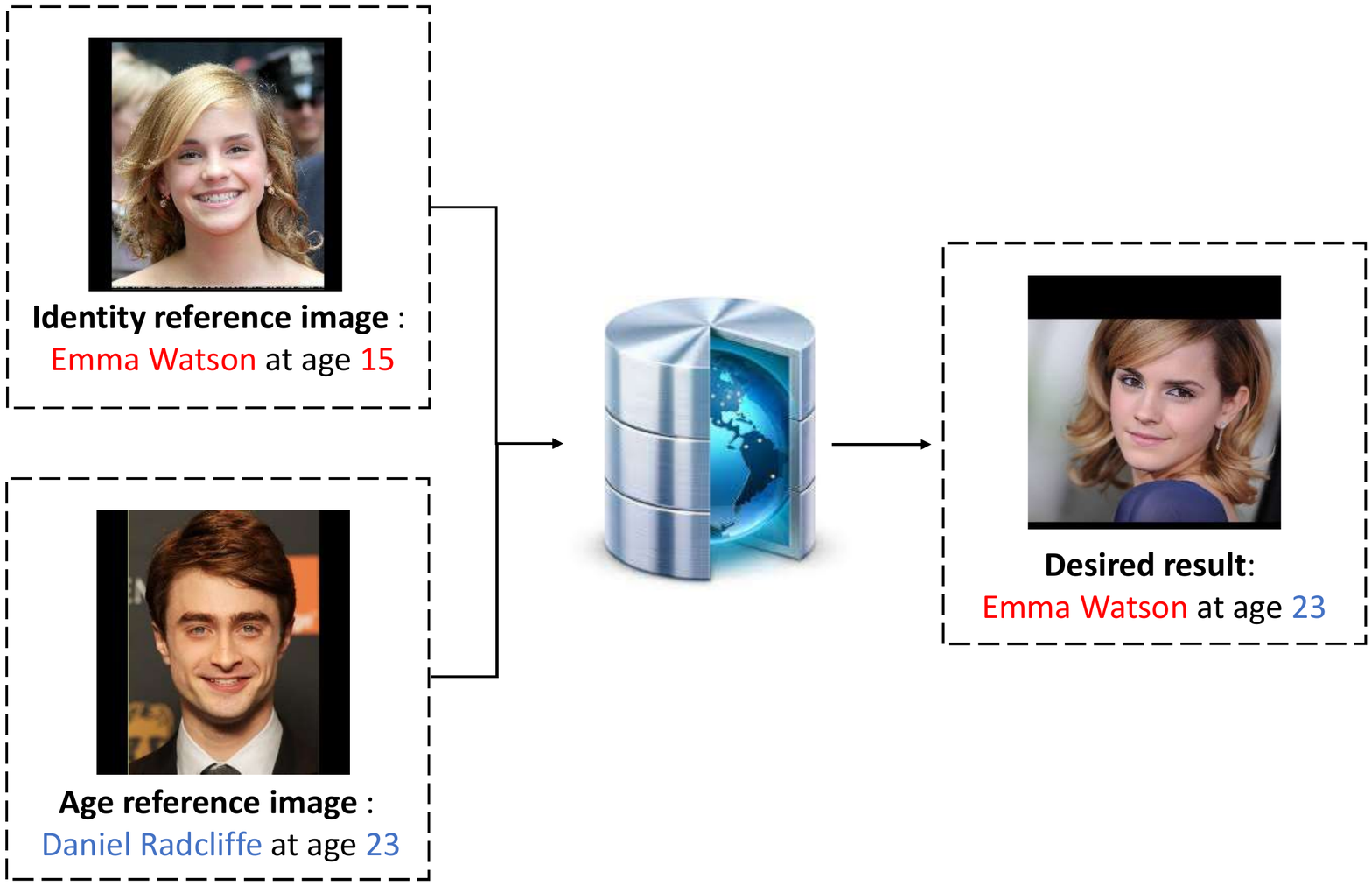}\\
		\caption{Dual-reference face retrieval framework: The system retrieves the images which is not only of the same identity in the identity reference image, but also at the similar age reflected in the age reference image.}
	\end{subfigure}
%	\subfigure[Conventional face retrieval framework: Firstly all the images similar to \textit{Emma Watson} are selected, then an external age estimator is employed to select images at the desired age, which is given as a numeral.]{\includegraphics[width=0.5\textwidth]{figures/conventional_fr.pdf}}\quad
%	\subfigure[Dual-reference face retrieval framework: The system retrieves the images which is not only of the same identity in the identity reference image, but also at the similar age reflected in the age reference image]{\includegraphics[width=0.5\textwidth]{figures/drfr.pdf} }
%	\caption{Comparison between conventional face retrieval framework and our proposed dual-reference face retrieval framework}
%	\label{fig:example}
	\caption{Comparison between conventional face retrieval framework and our proposed dual-reference face retrieval framework}\label{fig:example}
\end{figure*}

Considering such a task -- \textit{retrieving Emma Watson's image at 23}, although it is not absolutely impossible to be solved by the conventional face retrieval frameworks, for example, one can achieve it by concatenating an age estimation system at the end of the traditional face retrieval system to select the right images as illustrated in Figure.~\ref{fig:example}.(a), there exist many drawbacks in such a hierarchical framework. One of them is that using a single numeral is not capable to describe the human perceptions of the age, because the human performance on age estimation is with a large mean absolute error (MAE) as well as a large variance \cite{han2013age}, which means generally a human prefers to guess the age within a range rather than a certain numeral. Also, for humans, it is easier to estimate someones' age by comparing with age-known faces than directly assigning a facial image to a numeral\cite{chang2011ordinal}.
As the old saying goes, `One look is worth a thousand words', the problem of \textit{retrieving Emma Watson's images} is better solved by inputting one Emma Watson's picture and telling the machine: \textit{retrieving the images of the one shown in the input}. Since a numeral is not representative enough to describe a person's age, and in some scenarios, we do not even care about the certain age but the similarities in term of the age, what if we use an image to represent the target age? In this paper, we propose a novel face retrieval framework as shown in Figure.~\ref{fig:example}.(b), in which an age reference image besides the identity reference image is inputted to reflect the target age . We refer to the proposed framework as dual-reference face retrieval (DRFR). In the DRFR, the problem of \textit{retrieving Emma Watson's image at 23} is turned into \textit{retrieving the images of the one shown in the first input, and in the similar age reflected in the second input}.

In DRFR, the raw images are first projected onto a joint manifold, which preserves both the age and identity locality. Subsequently, as the age and identity are supposed to be measured differently on the joint manifold, a similarity metric for each is exploited and optimized via our proposed quartet-based model shown in Figure.~\ref{fig:quartet_model}. The final retrieval is conducted on the learned metric. 

The contributions of this paper mainly lie in the following three aspects:
\begin{enumerate}[1)]
	\item The task: retrieving someone's image at some age is an emerging task as more and more precise retrieval is required due to the explosive web images.
	\item The model: a joint manifold of identity and age is exploited in this paper, it simultaneously preserves the localities of these two aspects. Besides, a novel quartet-based model coordinated with two Mahalanobis distances is proposed to measure the similarities between the image pairs.
	\item The framework: our proposed DRFR task can be abstracted into a high-level task --- dual reference/query retrieval, which might lead to an emerging research direction. The existing retrieval methods generally take a single query or multiple queries indicating the same semantic information, while in our dual-reference framework, more than one semantic information can be taken into consideration.
\end{enumerate}

The remainder of the paper is organized as follows: we review the works related to the proposed task in Section 2; our proposal is outlined in detail in Section 3; in Section 4, we discuss the experimental results, and we provide a short conclusion in Section 5.

%------------------------------- End of Introduction ------------------------------%
%------------------------------- Related works ------------------------------------%
\section{Related Work}
To the best of our knowledge, the task of the dual reference face retrieval has never been raised in the literature, and there are no similar existing works, thus we review related works in the areas of face retrieval and age estimation, focusing on those papers which explore facial feature representation, age variation capturing and similarity metric learning. 

\textbf{Facial Feature Representation} A broad array of research \cite{luo2017robust,ou2014robust} has been completed on facial feature representation. As facial features extracting is not the core part of our framework, we just give a rough review here. For a comprehensive review, we refer our readers to \cite{bagherian2008facial}. Early works mainly take heuristic features such as Gabor \cite{liu2002gabor}, HOG \cite{dalal2005histograms}, LBP \cite{ahonen2006face} or their extensions. However, designing hand-crafted features is a trial and error process which is less than adequate for our purpose.
Another branch of research regarding facial features is based on utilizing deep learning. For example, \cite{taigman2014deepface} employed a nine-layer deep neural network to extract facial features for face verification and \cite{NIPS2014_5416} proposed a precisely designed deep convolutional networks for joint face identification-Verification.

\textbf{Age Variation Capturing} Age variation capturing is rarely considered in conventional face retrieval approaches because in most works to date, features are required to be age-invariant. In contrast, we are seeking a facial representation that embeds both identity and age information. 
Approaches capturing age variation can primarily be found in age estimation literature. The earliest approach of age estimation based on facial images dates back to 1994, \cite{kwon1994age} uses geometric features, in which the ratios between different measurements of facial landmarks (\emph{e.g.} eyes, chin, nose, mouth, etc.) are calculated to classify the individual into three age groups, namely \textit{infants}, \textit{young adults} and \textit{senior adults}. Unfortunately, it suffers in distinguishing young and old adults as both the shape and texture of the face change during aging \cite{suo2007multi}. To overcome the drawbacks of geometric features, the Active Appearance Model (AAM) is proposed in \cite{cootes2001active}. AAM is able to simultaneously capture the shape and texture information of face images. Considering the temporal characteristics of human aging, Aging Pattern Subspace \cite{geng2008facial} treats a serial of a person's images as an aging pattern thus the information during aging process is embedded.

\textbf{Similarity Metric Learning} Once the proper facial image representation is selected, the retrieval is conducted based on the similarity measurements. There are many works \cite{zhao2017unconstrained,guo2017robust,zheng2016learning,guo2017zero} focusing on the similarity metric learning. \cite{hadsell2006dimensionality} induced a contrastive loss to ensure that the neighbors are pulled together while the non-neighbors are pushed apart on the learned metric. Different with the contrastive loss that only considers pairwise examples at a time, \cite{wang2014learning} and \cite{schroff2015facenet} proposed the triplet loss, which minimizes the $L_2$-distance between an anchor and a positive sample, both of which belong to the same instance, and maximizes the distance between the anchor and a negative sample. However, the traditional triplet-loss may lead to a large intra-class variation during testing. \cite{chen2017beyond} added a fourth sample in the triplet to enlarge the inter-class variation thus reducing the intra-class variation.
%------------------------------- End of Related works -----------------------------%

%------------------------------- Methods ------------------------------------------%
\section{Dual-Reference Face Retrieval}

For convenience, define $I_i^m$ as an image of the individual with identity $i$ at age $m$.  Input an image pair $(I_i^m,I_j^n)$, where $i$ is the target identity and $n$ is the objective age, thus our required output is $I_i^n$. As discussed, DRFR consists of two stages. Firstly, a mapping function is learned to project the raw images onto a joint manifold. Subsequently, to measure the similarity between each pair of images, the two metrics are learned on the low-dimensional space, based on a quartet model. We devote the rest of this section to outlining these two stages.   
\subsection{Joint Manifold}
\begin{figure}
	\includegraphics[width=1\linewidth]{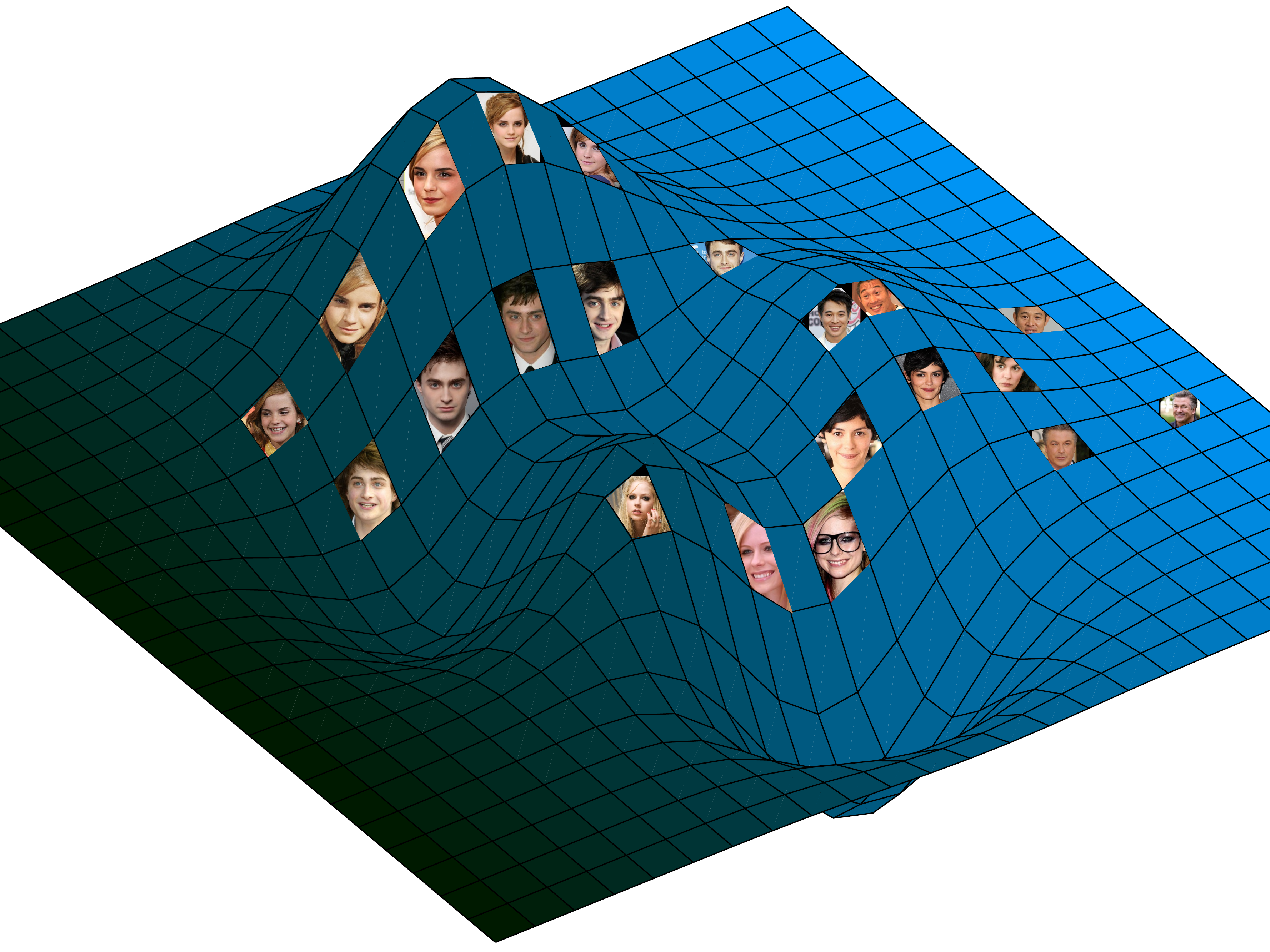}\\
	\caption{An illustration of the joint manifold of age and identity.}\label{fig:joint-manifold}
\end{figure}
A face image with $\mathcal{D}$-dimensional feature representation can be considered as a point in the $\mathcal{D}$-dimensional space containing rich information such as age, gender, race, identity. Manifold learning is first proposed in \cite{roweis2000nonlinear}, in which they believe that the high-dimensional data is sampled from a smooth low-dimensional manifold. Thus it is natural that information from a facial image can be represented within low-dimensional manifolds embedded in a high-dimensional image space \cite{he2005face}. Many applications \cite{zheng2016hetero,zheng2013semi} already utilize low-dimensional manifolds to embed human face images, such as face recognition and age estimation. However, our proposed joint manifold as illustrated in Figure.~\ref{fig:joint-manifold} is very different; instead of treating the age and identity as two separate degrees of freedom in a single manifold, with the assumption that the age and identity are both manifolds sampled from a higher-dimensional manifold. 

Let $\mathcal{X}$ be the original representation of the raw images and $\mathcal{Y}$ be the low-dimensional joint manifold, define the mapping function of the joint manifold to be $f: \mathcal{X} \rightarrow \mathcal{Y} $. Since both the locality of the age and identity can be represented as matrices, let $S$ denote the set of all such similarity matrices. Specifically, the matrix $S^n \in S$ reflects the similarity among all the individuals' images at age $n$; similarly, $S_i$ denotes the similarity over those images belonging to an individual with identity $i$ across all ages. The desired properties of $f$ are discussed below. 

\textbf{Preserving Locality of Individual Space}
We first calculate the similarity matrix $S^n$. In detail, among all the images at age $n$, if two images are nearby in original feature space $\mathcal{X}$, we mark the similarity as $\exp\left(-\frac{\parallel x_i^n-x_j^n\parallel^2_2}{t}\right)$, where $x_i^n \in \mathcal{X}$ is the original feature representation of image $I_i^n$ and $\parallel \cdot \parallel^2_2$ is the $l2$-norm, otherwise their similarity is $0$. Thus the similarity matrix $S^n$ under age $n$ is calculated as:
\begin{equation}
S^n(x_i^n,x_j^n)=
\begin{cases*} 
\exp\left(-\frac{\parallel x_i^n-x_j^n\parallel^2_2}{t}\right) & if $x_j^n \in \mathcal{N}(x_i^n)$\com \\
0                                &otherwise\com
\end{cases*} 
\end{equation}
where $\mathcal{N}(x_i^n)$ denotes the neighbors of $x_i^n$. To preserve the locality, we require the nearby points in $\mathcal{X}$ to remain close to each other after being embedded into $\mathcal{Y} = f(\mathcal{X})$, thus we optimize the function:
\begin{equation}\label{eqn:s1}
\underset{f}\min \underset{n}\sum\underset{i,j}\sum\parallel f(x_i^n)-f(x_j^n)\parallel ^2_2S^n(x_i^n,x_j^n)\fs
\end{equation}

\textbf{Preserving Locality of Age Space}
Similarly, to calculate the age similarity matrix $S_i$, we gather all the images of the individual $i$, and assign $\exp\left(-\frac{\parallel x_i^n-x_i^m\parallel ^2_2}{t}\right)$ as the similarity if $m-n$ is below a threshold $\varepsilon$, otherwise the similarity is $0$:
\begin{equation}
S_i(x_i^n,x_i^m)=
\begin{cases*} 
\exp\left(-\frac{\parallel x_i^n-x_i^m\parallel ^2_2}{t}\right) & if $|m-n|< \varepsilon$, \\
0                                &otherwise.
\end{cases*} 
\end{equation}
To preserve the local smoothness, we optimize the function:
\begin{equation}\label{eqn:s2}
\underset{f}\min \underset{i}\sum\underset{m,n}\sum\parallel f(x_i^n)-f(x_i^m)\parallel ^2_2S_i(x_i^n,x_i^m)\fs
\end{equation}

\subsection{Similarity Metric Learning Based on a Quartet Model}
\begin{figure}
	\begin{center}
		\includegraphics[width=0.6\linewidth]{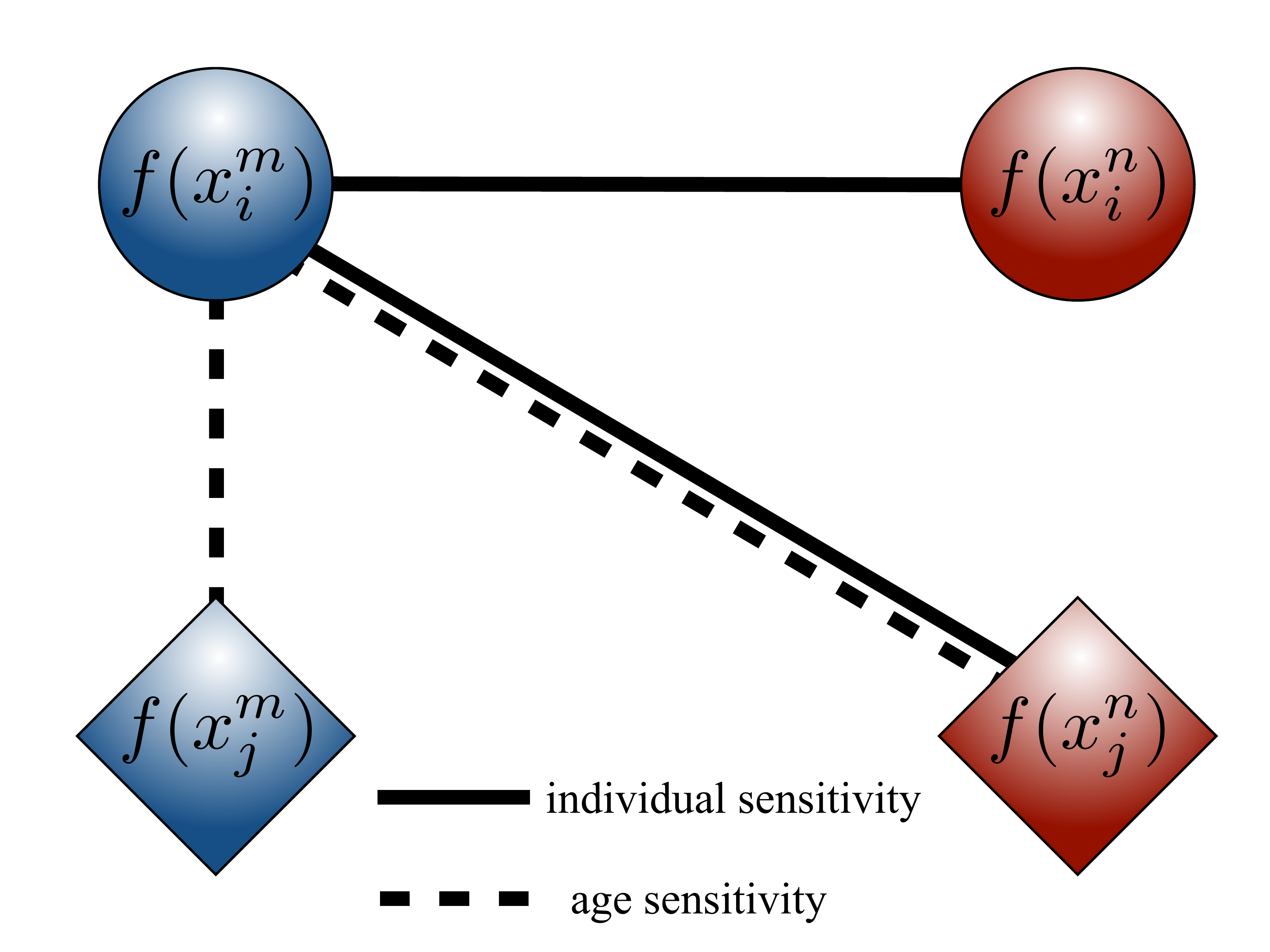}
	\end{center}
	\caption{An illustration of our proposed quartet model. The blue symbols indicate the embedded points of images at age $m$ and the red ones stand for those at age $n$. The circle symbols represent the embedded points of images of individual $i$ while the diamond ones stand for those of individual $j$. The lengths of the lines connecting any two symbols can be regarded as the distance between the corresponding embedded points. Thus in any triangle in the quartet sample, the length of its hypotenuse is larger than that of its leg.}\label{fig:quartet_model}
\end{figure}
After both the original age and identity spaces are mapped onto a joint manifold, different measurements should be taken to obtain the similarity of the two aspects. In this paper, two similarity metrics are learned based on a novel quartet model, which is a graph with 4 vertices as shown in Figure.~\ref{fig:quartet_model}. The vertices sets $V=\{f(x_i^m),f(x_i^n),f(x_j^m),f(x_j^n)\}$ are the embedded points of $\{(x_i^m),(x_i^n),(x_j^m),(x_j^n)\}$, and the edges are defined as the distance between each embedded point. 
We use $\Phi(\cdot,\cdot)$ to denote the difference measurement function whereby the smaller $\Phi(\cdot,\cdot)$ is, the more similar the two images are. In the following of this subsection, the properties of the desired metrics are introduced.

\textbf{Individual Metric}
Considering two image pairs $(x_i^m,x_i^n)$ and $(x_j^m,x_i^n)$, which are shown in the quartet model in Figure.~\ref{fig:quartet_model}, it is very clear that on the individual metric, the distance between $x_i^m$ and $x_i^n$ is smaller than that between $x_i^m$ and $x_j^n$, because these two pairs of images both have the age gap $m-n$ while the first image pair $(x_i^m,x_i^n)$ belongs to the same individual $i$. Mathematically, there is:
\begin{equation}\label{eqds1}
\begin{split}
&\Phi_{ind}(f(x_i^m),f(x_i^n)) < \Phi_{ind}(f(x_i^m),f(x_j^n)) \\
&\forall (i,j,m,n)\com
\end{split}
\end{equation}
where $\Phi_{ind}$ measures the individual difference between any pair of images.

Additionally, the distances between image pair $(x_i^m,x_j^m)$ and $(x_i^m,x_j^n)$ are supposed to be similar because the individual metric is uncorrelated with the age, which can be written as:
\begin{equation}\label{eqs1}
\begin{split}
&\Phi_{ind}(f(x_i^m),f(x_j^m)) = \Phi_{ind}(f(x_i^m),f(x_j^n)) \\
&\forall (i,j,m,n)\com
\end{split}
\end{equation}

\textbf{Age Metric}
Similarly on the age metric, the distance between image pair $(x_i^m,x_j^m)$ is smaller than that between $(x_i^m,x_j^n)$, and the distances are close if the age gap within each image pair is same. Thus we have:
\begin{equation}\label{eqds2}
\begin{split}
&\Phi_{age}(f(x_i^m),f(x_j^m)) < \Phi_{age}(f(x_i^m),f(x_j^n)) \\
&\forall (i,j,m,n)\com
\end{split}
\end{equation}
\begin{equation}\label{eqs2}
\begin{split}
&\Phi_{age}(f(x_i^m),f(x_j^n)) = \Phi_{age}(f(x_i^m),f(x_i^n)) \\
&\forall (i,j,m,n)\com
\end{split}
\end{equation}
where $\Phi_{age}$ measures the age difference between any pair of images.

\textbf{Quartet Loss}
To obtain the discussed characteristics of the individual and age metrics, a loss function which maximize the margin between the distances in Eq.~\ref{eqds1} and Eq.~\ref{eqds2}, and meanwhile minimize the margin between the distances in Eq.~\ref{eqs1} and Eq.~\ref{eqs2} is designed. For convenience, we first define $d$ as the distance of two images embedded in the joint manifold $\mathcal{Y}$: $d_{ij}^{mn} = f(x_i^m)-f(x_j^n)$ and take the Mahalanobis distance as the distance measurement. Thus the $\Phi(\cdot,\cdot)$ can be written as:
\begin{equation}
\begin{split}
&\Phi_{age}(f(x_i^m),f(x_j^n))={d_{ij}^{mn}}^\top\mathbf{M}_{\text{age}}{d_{ij}^{mn}}\com \\
&\Phi_{ind}(f(x_i^m),f(x_j^n))={d_{ij}^{mn}}^\top\mathbf{M}_{\text{ind}}{d_{ij}^{mn}}\com
\end{split}
\end{equation}
where $\mathbf{M}_{\text{age}}$ and $\mathbf{M}_{\text{ind}}$ are the Mahalanobis matrices. 
To maximize the margin, the hinge loss function:
\begin{equation}\label{eq10}
H(y)=\max(0,\delta-y)
\end{equation} is employed.

Thereby for a quartet sample indexed by $(i,j,m,n)$, the loss $\mathcal{L}_{ij}^{mn}$ can be defined as:

%\begin{equation}
\begin{align*}
\mathcal{L}_{ij}^{mn} = & H({d_{ij}^{mn}}^\top\mathbf{M}_{\text{age}}{d_{ij}^{mn}}-{d_{ij}^{mm}}^\top\mathbf{M}_{\text{age}}{d_{ij}^{mm}})\\
+ & H({d_{ij}^{mn}}^\top\mathbf{M}_{\text{ind}}{d_{ij}^{mn}}-{d_{ii}^{mn}}^\top\mathbf{M}_{\text{ind}}{d_{ii}^{mn}})\\
+ & ||{d_{ij}^{mn}}^\top\mathbf{M}_{\text{age}}{d_{ij}^{mn}}-{d_{ii}^{mn}}^\top\mathbf{M}_{\text{age}}{d_{ii}^{mn}}||^2_2 \\
+ & ||{d_{ij}^{mn}}^\top\mathbf{M}_{\text{ind}}{d_{ij}^{mn}}-{d_{ij}^{mm}}^\top\mathbf{M}_{\text{ind}}{d_{ij}^{mm}}||^2_2.
\end{align*}
%\end{equation}
And the loss over the whole training set is
\begin{equation} \label{eq8}
\mathcal{L}=
\underset{i,j,m,n} \sum L_{ij}^{mn} \fs
\end{equation}

\subsection{Optimization}
Considering the loss function $\mathcal{L}$ and the joint manifold as the regularization term, the overall objective function is:
\begin{equation}\label{eqn:objective}
\begin{split}
\mathcal{J}=\mathcal{L}+&\underset{n}\sum\underset{i,j}\sum\parallel d_{ij}^{nn}\parallel ^2_2S^n(x_i^n,x_j^n)\\
+&\underset{i}\sum\underset{m,n}\sum\parallel d_{ii}^{mn}\parallel ^2_2S_i(x_i^m,x_i^n)\com \\
\text{s.t.}&~\mathbf{M}_{\text{ind}} \succeq 0, \mathbf{M}_{\text{age}} \succeq 0\com
\end{split}
\end{equation} 
where $\mathbf{M} \succeq 0$ implies that $\mathbf{M}$ is a semi-definite positive matrix, thus pseudometrics are allowed.

As both the Mahalanobis matrices $\mathbf{M}_{\text{age}}$ and $\mathbf{M}_{\text{ind}}$ as well as the embedding function $f$ need to be learned in Eq.~\ref{eqn:objective}, we employ a deep network to optimize them jointly. The architecture of the proposed network is discussed in the following sections.

\begin{figure}
	\begin{center}
		\includegraphics[width=1\linewidth]{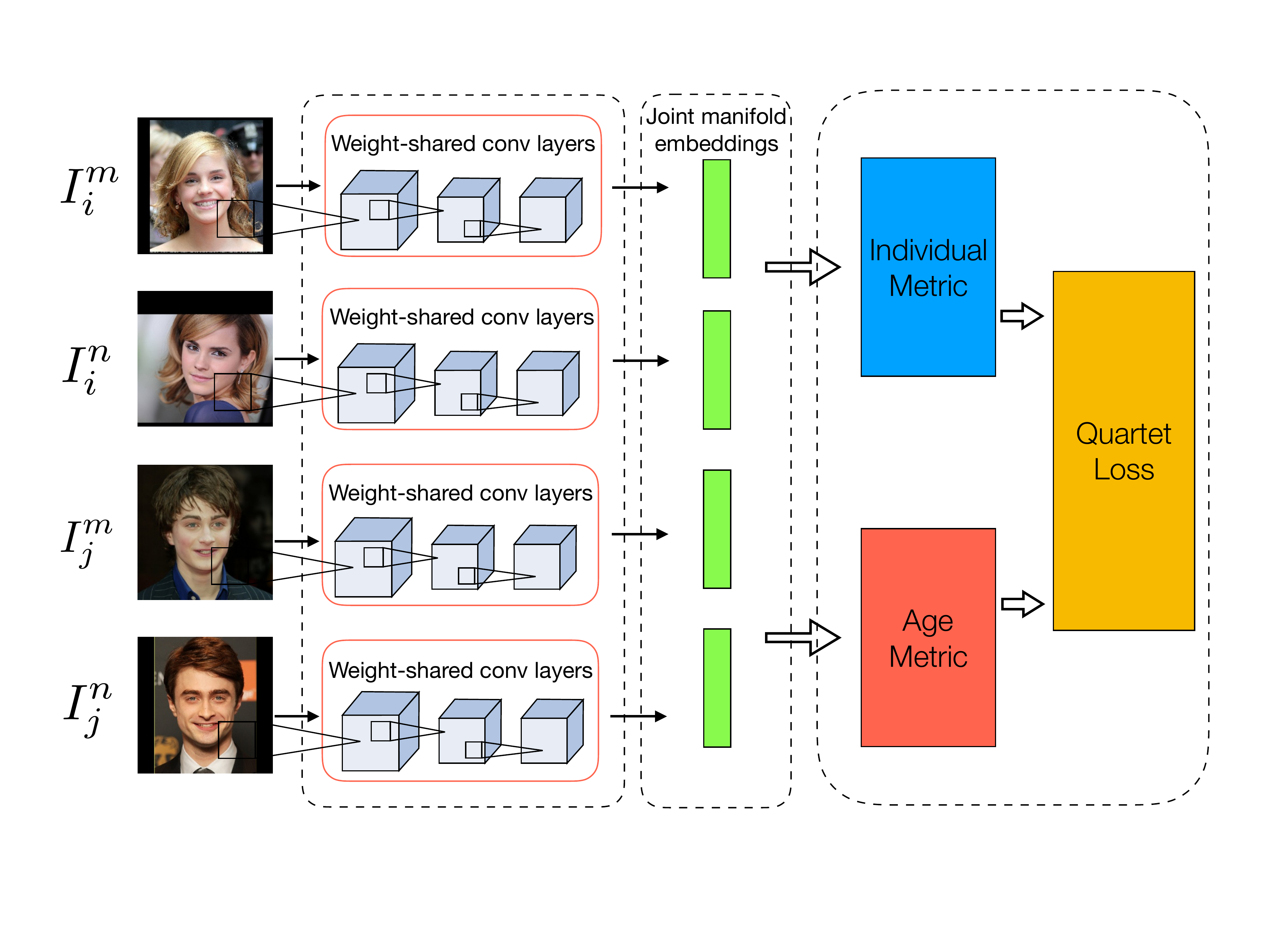}
	\end{center}
	\caption{The architecture of our proposed deep network. The network takes quartet samples as input, and the joint manifold embeddings are obtained after the images are forward propagated through four weight-shared convolutional layers. Subsequently, the distances between embedded images on the joint manifold are measured by two independent metrics -- individual metric(blue) and age metric(red). Finally the distances are feed to the last layer to optimize the quartet loss.}\label{fig:arch}
\end{figure}

\textbf{Deep Network Architecture}
Our quartet-based network architecture is shown in Figure.~\ref{fig:arch}, which jointly optimizes the manifold embedding function $f$ and the two Mahalanobis matrices. The network takes quartet samples as input. Each quartet sample contains an image set $Q=\{x_i^m,x_i^n,x_j^m,x_j^n\}$, which are the images of the person $i$ and $j$ at his $m$ and $n$ age stage. The images are firstly passed through a weight-shared convolutional layers, which can extract prolific and robust age and identity information from a facial image while preserving the locality. The deep convolutional network takes the joint manifold cost as the loss function. Subsequently, the distance between the outputs of the deep architecture, for example, $f(x_i^m)$ and $f(x_i^n)$ are measured via two independent metrics, which are namely, age metric and individual metric. With the distances between each image pairs, the quartet loss are thus optimized and the gradients are back-propagated to update the $\mathbf{M}$.

\textbf{Deep Convolutional Layer}
In our model, the deep convolution layer is trained to explore the joint manifold of the age and identity. As discussed that the joint manifold is supposed to keep the locality structure, thus the Eq.~\ref{eqn:s1} and Eq.~\ref{eqn:s2} are taken as the joint manifold cost. In the experiment, we first compute the similarity matrix across the whole dataset while for each input batch, only the involved locality constraints need to be satisfied during training, which leads to a great computation saving. As a fact, the linear embedding can already reflect the joint manifold, however we employ the deep learning for a better performance. 
\begin{figure*}
	\begin{center}
		\includegraphics[width=0.7\linewidth]{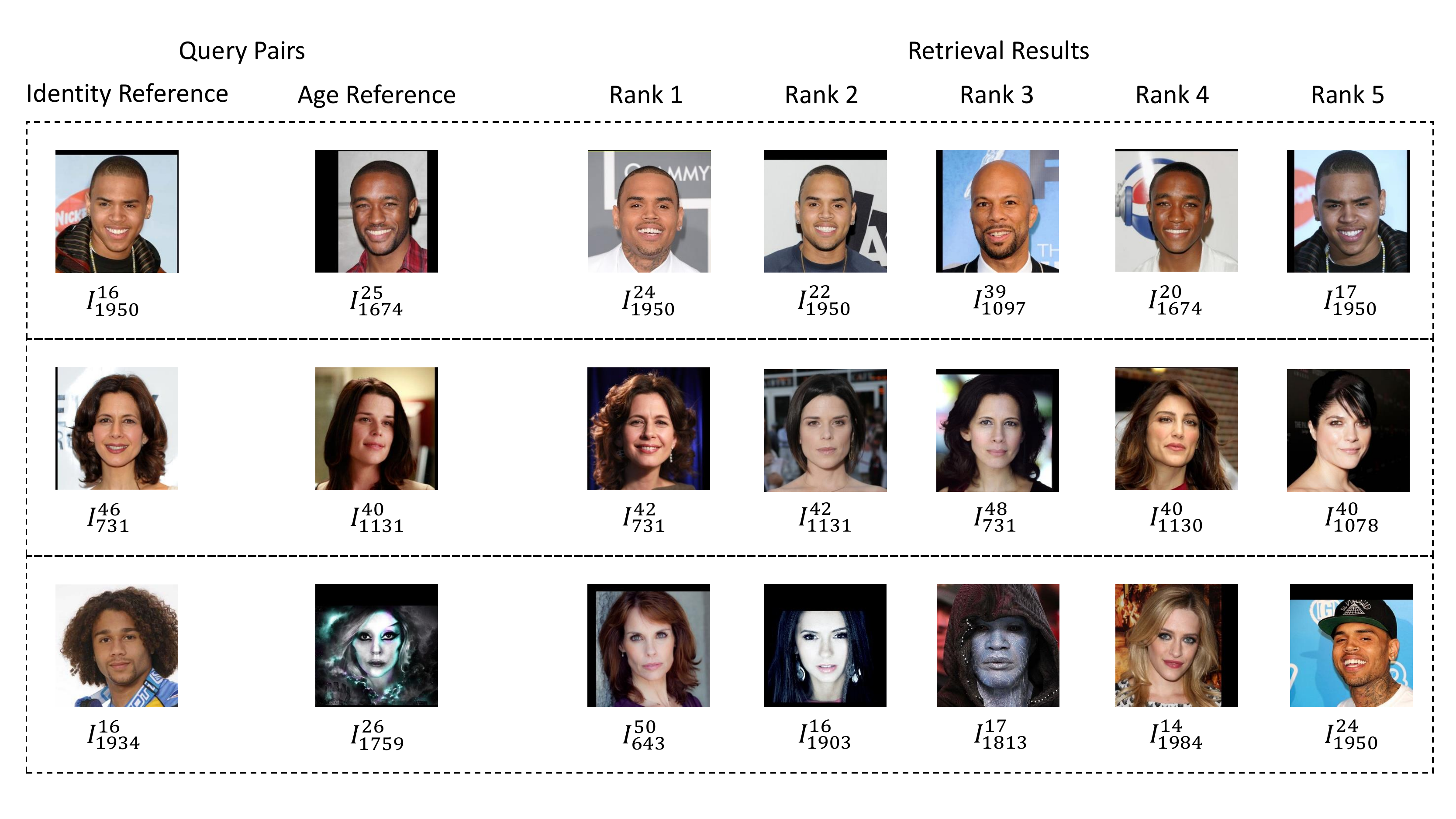}
	\end{center}
	\caption{Experimental results on CACD dataset. The first row and second row are selected two convincing retrieval results and the third row is a picked bad retrieval example. However, the failure shown here is because that the age reference image contains too much noisy and even a human cannot correctly figure out the age of the subject, thereby such noisy data influenced the similarity measurement both on the age metric and the individual metric.}
	\label{fig:long}
\end{figure*}

\textbf{Individual Metric and Age Metric Learning}
At the end of the deep architecture module, the facial images are represented by a $d$-dimensional feature. To measure the distances between each image, we introduce two Mahalanobis matrices $\mathbf{M}_{\text{age}}$ and $\mathbf{M}_{\text{ind}}$. Since Mahalanbis matrices are semi-definite positive, $\mathbf{M}$ can be factorized as $\mathbf{M} = L^{\top}L$. In other words, to learn the individual metric and age metric is equally to learn two projections $L_{\text{ind}}$ and $L_{\text{age}}$ as:
\begin{equation}
\begin{split}
\Phi(f(x_i^m),f(x_j^n)) & ={d_{ij}^{mn}}^\top\mathbf{M}{d_{ij}^{mn}}\\
& = ||Lf(x_i^m)-Lf(x_j^n)||_2^2
\end{split}
\end{equation}
In our architecture, the two metrics layer are inner product layers with independent weights. The eucledean distance in the projected space is the corresponding Mahalanbis distance. It is not hard to update the matrix $L$ via the loss function Eq.~\ref{eqn:objective} while how to ensure $\mathbf{M}$ being semi-positive is a problem. Inspired by \cite{Shai2004}, we take a trick when updating on $L$ happens. After $L$ is updated by the network, we check all the eigenvalue of the matrix $L$ and change the most negative eigenvalue to zero and then update $L$ again to make it closer to a semi-positive matrix.
%------------------------------- End of Methods -----------------------------------%
%-------------------------------- Experiments -------------------------------------%
\section{Experiment}
As the dual-reference face retrieval is a newly explored task, there are few datasets where each individual's images have a wide age range. However, we emphasize that the scarcity of suitable datasets does not mean the task is unnecessary. On the contrary, it reflects the fact that using dual reference images to indicate multiple semantic information is reasonable when merged by the huge volume of unlabelled online images. 

In the experiment, we evaluate our DRFR on three face recognition and age estimation datasets: Cross-Age Celebrity Dataset(CACD) \cite{chen2014cross}, FGNet \cite{lanitis2002fg}, and MORPH \cite{ricanek2006morph}. The statistics of these datasets are shown in Table.~\ref{table_distribution}. As the CACD contains the most images among the three, we trained our deep neural network and conducted our main experiments on the CACD. Apart from that, we evaluated the robustness of our joint manifold model on FGNet and performed the cross-dataset validation on the MORPH. 

\begin{table}[h]
	\begin{center}
		\begin{tabular}{|l|c|c|c|c|}
			\hline
			Dataset & \multicolumn{1}{l|}{Images} & \multicolumn{1}{l|}{Subjects} & \multicolumn{1}{l|}{Images/sub.} & \multicolumn{1}{l|}{Age gap} \\ \hline
			CACD & 163446 & 2000 & 81.7 & 0-9 \\ \hline
			FGNet & 1002 & 82 & 12.2 & 0-45 \\ \hline
			MORPH & 55134 & 13618 & 4.1 & 0-5 \\ \hline
		\end{tabular}
	\end{center}
	\caption{Statistics of the Datasets }
	\label{table_distribution}
\end{table}

\subsection{Experiment on CACD}
\textbf{Settings} The Cross-Age Celebrity Dataset is collected for the cross-age face retrieval task in \cite{chen2014cross}, and it contains $163446$ images from $2000$ celebrities with the age ranging from $16$ to $62$. The large-scale data with high age variations provides the DRFR ideal experimental conditions. However, it is noteworthy that although the age ranges from $16$ to $62$, the maximum age gap for each celebrity is 9 years old, as all the collected images are taken from 2003 to 2014. In detail, the age gaps are stepping at 1 year old from $(14-23)$ to $(53-62)$, thus there are 40 age gaps in total. On average, each age gap contains $4000$ images of 50 celebrities. Following the settings in \cite{chen2014cross}, we take $60\%$ data as training data and the remaining for the test. The training data is picked uniformly from each age gap to ensure all the age gaps are covered. For the test data, as there are $8$ different images for each celebrity at each age in average, we further split the test data into $8$ subsets for the following evaluation.
To train our deep network on DRFR, the weights of two Mahalanobis matrices were initialized as identity matrices. For the hyper-parameters, we set the $\varepsilon$ in Eq.~\ref{eq8} as $5$ to calculate the similarity matrix set $S$, and the embeddings' size on the joint manifold is set as $128$. The triplet selection scheme can heavily impact the convergence speed of the network training, so does the quartet samples selection. An effective triplet selection can avoid poor training and reduce the influences caused by the mislabelled data, we employed an online quartet selection protocol which is inspired by \cite{chen2017beyond}. During training, the images of an entire mini batch are firstly propagated forward to extract the embeddings with the current model, then those quartets which violate the average margin in this mini batch will be selected to train the network.

\textbf{Evaluation Metrics and Comparison} As DRFR can be regarded as a fine-grained retrieval, we use the top-$K$ retrieval accuracy\cite{wang2014learning} as the evaluation metric. Since there are no works on this task in the literature before, we combined the existing face retrieval approaches with the age estimation methods to form a hierarchical framework and made the comparison. In the combined hierarchical framework, the face retrieval was first conducted regarding the first reference image as the query. Subsequently, we estimated the age of the second reference image and the top 100 candidate images from the face retrieval session. Finally these 100 images are ranked according to the estimated ages. For the facial representation, we choose eigenfaces, LBP, CARC\cite{chen2014cross}, which encodes the images with a set of celebrities, and the deep learning feature extracted from the FaceNet\cite{schroff2015facenet}. In the following age estimation, we selected support vector regression (SVR) as well as canonical-correlation analysis (CCA). For the FaceNet feature, we used the same training data as our DRFR's. 

\textbf{Results}
We conducted DRFR and the 8 hierarchical methods on the 8 testing subsets and compute the average top-$K$ retrieval accuracy. The results are shown in Table.~\ref{experiment_result}. It shows that when the $K$ is small(less than 6), our proposed DRFR outperformed the other 4 three methods. It is interesting to note that when the allowed output image increases, the accuracy of CARC+CCA is slightly higher than ours. The reason is that CACD is the original dataset which CARC designed, and in our settings, each subset only contains approximately 10 images for each subject, it is reasonable for a high accuracy if the face retrieval system can retrieve all the images of the correct identity.    

\begin{figure}
	\begin{center}
		\includegraphics[width=0.8\linewidth]{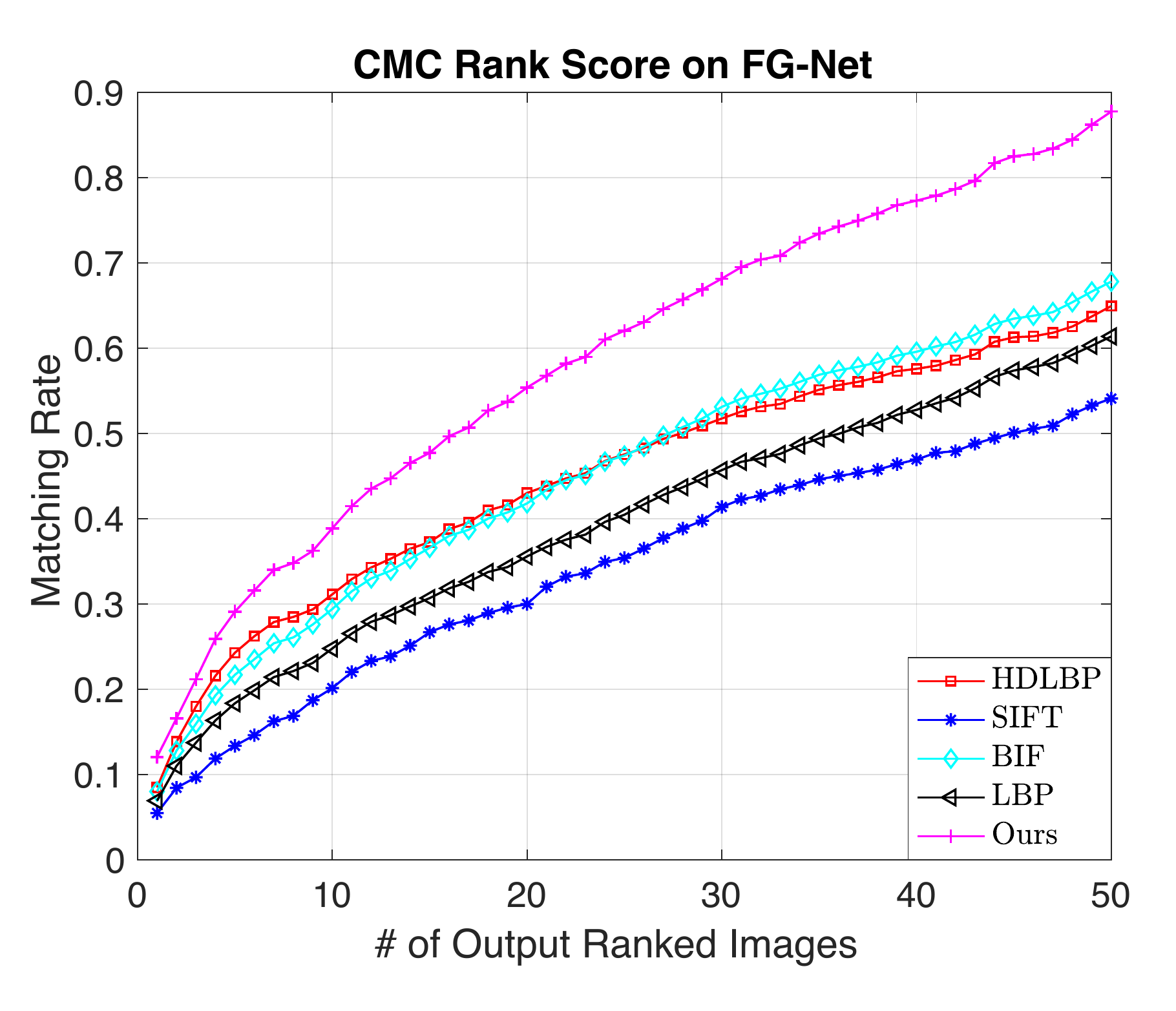}
	\end{center}
	\caption{The results of the experiment on FGNet.}
	\label{fig:result}
\end{figure}

\begin{table*}
	\begin{center}
		\begin{tabular}{|c|r|r|r|r|r|r|r|}
			\hline
			\multicolumn{1}{|l|}{Accuracy\% @ top-$K$} & \multicolumn{1}{c|}{$K$=1} & \multicolumn{1}{c|}{$K$=2} & \multicolumn{1}{c|}{$K$=3} &\multicolumn{1}{c|}{$K$=4} &\multicolumn{1}{c|}{$K$=5} & \multicolumn{1}{c|}{$K$=8} & \multicolumn{1}{c|}{$K$=10} \\ \hline
			eigenfaces+SVR     & 14.43 & 17.25 &17.42 &17.87 & 18.5  & 19.10 & 19.20 \\
			eigenfaces+CCA     & 14.97 & 17.73 &18.21 &18.53 & 18.71 & 19.24 & 19.35 \\
			LBP+SVR            & 17.58 & 20.32 &20.86 &21.52 & 21.85 & 23.45 & 24.53 \\
			LBP+CCA            & 17.98 & 21.44 &22.13 &22.13 & 22.22 & 24.78 & 25.71 \\
			CARC+SVR           & 18.34 & 22.45 &23.02 &23.64 & 24.30 & 25.70 & 26.20 \\
			CARC+CCA           & 18.57 & 22.25 &23.50 &23.85 & 24.50 & \textbf{26.12} & \textbf{26.42} \\
			FaceNet+SVR        & 19.76 & 23.20 &23.33 &23.77 & 23.64 & 24.70 & 26.33 \\
			FaceNet+CCA        & 19.63 & 23.48 &24.12 &24.37 & 24.54 & 25.38 & 26.40 \\ \hline
			DRFR(Ours)         & \textbf{20.67} & \textbf{23.75} &\textbf{24.33} &\textbf{24.87} & \textbf{24.90} & 25.80 & 26.23 \\ \hline 
		\end{tabular}
	\end{center}
	\caption{Experimental results on CACD dataset.}
	\label{experiment_result}
\end{table*}

\subsection{Experiment on FGNet}
\textbf{Settings} FGNet dataset consists of 1002 images of 82 subjects in total. As it is tiny while has high age variations, we conducted experiments using different features on it to evaluate performance of the joint manifold embedding function $f$ of our proposed framework. Similar to the experiment setting on CACD, we split FGNet into training and test set, avoiding the situation that the same subject shows in both sets. The training set contains $60\%$ images while the rest is left for the testing.

\textbf{Comparison with Linear Embedding Method}
To evaluate the robustness of our joint manifold embedding, we extracted the embeddings, which is shown as green stripes in Figure.~\ref{fig:arch}, from the model trained in above CACD experiments. And we chose four other feature descriptors, which includes: LBP, BIF, SIFT and HDLBP, to make the comparison. To get the corresponding embeddings of these hand-crafted features, we employed PCA as the embedding technique, whose projection matrix is denoted as $W$. Subsequently, the age and individual metrics $\mathbf{M}_{\text{age}}$ and $\mathbf{M}_{\text{ind}}$ were trained for each embedding based on the quartet loss. And finally the retrieval was conducted on the learned metrics. 

\textbf{Results} Figure.~\ref{fig:result} shows the results of our experiments on FGNet. It can be seen that the CMC rank score of our joint manifold outperforms others. Since the 
$\mathbf{M}_{\text{age}}$ and $\mathbf{M}_{\text{ind}}$ are learned with respect to each embedding, we can draw the conclusion that: firstly, the joint manifold embedding function trained on CACD has a robust generalization. Secondly, the proposed joint manifold preserves more information of the age and identity locality.

\subsection{Cross-Dataset Validation on MORPH}
The MORPH dataset has 55134 images of 13618 subjects. Though both the images and subjects are in big amount, the number of images for each subject is only 4.1, which is not sufficient to compromise the quartet samples for training. Thereby instead of training a new model, we conduct a cross-validation on MORPH. 
We used the model trained on the CACD dataset directly on the MORPH dataset and the results are shown in Table.~\ref{experiment_morph}. It is shown that the results are very close to those on CACD. One reason of the minor backward can be the divergence of the age distribution between MORPH and CACD. Another reason is that the images in MORPH are over-cropped and some parts of the forehead and the chin in the image are absence, while the images are all of the full faces in CACD.

\begin{table}
	\begin{center}
		\begin{tabular}{|c|r|r|r|r|r|r|r|}
			\hline
			\multicolumn{1}{|l|}{Acc\% @ top-$K$} & \multicolumn{1}{c|}{$K$=1} & \multicolumn{1}{c|}{$K$=3} &\multicolumn{1}{c|}{$K$=5} &  \multicolumn{1}{c|}{$K$=10} \\ \hline
			
			MORPH       & 18.26 & 20.81 &22.99 &23.17  \\ \hline
			CACD         & 20.67  &24.33  & 24.90  & 26.23 \\ \hline 
		\end{tabular}
	\end{center}
	\caption{Cross dataset validation on MORPH.}
	\label{experiment_morph}
\end{table}

\section{Conclusions and Future Work}
In this paper, we proposed a dual-reference face retrieval framework, which tackles the problem of retrieving a person's face image at a `given' age. In the proposed framework, the retrieval is conducted on a joint manifold and based on two similarity metrics.
We have systematically evaluated our approach on CACD, FGNet and MORPH, and the corresponding results show that the proposed approach achieves promising results on this new task and the framework is stable and robust. 

For the future work, a larger dataset with wider age range can be collected to further improve our algorithm. Also, the dual-reference retrieval framework can be extended to other retrieval tasks besides the face. 
%------------------------------End of Experiments ---------------------------------%

\bibliography{egbib}

\begin{thebibliography}{}

\bibitem[\protect\citeauthoryear{Ahonen, Hadid, and
  Pietikainen}{2006}]{ahonen2006face}
Ahonen, T.; Hadid, A.; and Pietikainen, M.
\newblock 2006.
\newblock Face description with local binary patterns: Application to face
  recognition.
\newblock {\em IEEE TPAMI} 28(12):2037--2041.

\bibitem[\protect\citeauthoryear{Bagherian and
  Rahmat}{2008}]{bagherian2008facial}
Bagherian, E., and Rahmat, R. W.~O.
\newblock 2008.
\newblock Facial feature extraction for face recognition: a review.
\newblock In {\em 2008 International Symposium on Information Technology},
  volume~2,  1--9.

\bibitem[\protect\citeauthoryear{Bhattacharjee \bgroup et al\mbox.\egroup
  }{2011}]{bhattacharjee2011construction}
Bhattacharjee, D.; Halder, S.; Nasipuri, M.; Basu, D.~K.; and Kundu, M.
\newblock 2011.
\newblock Construction of human faces from textual descriptions.
\newblock {\em Soft Computing} 15(3):429--447.

\bibitem[\protect\citeauthoryear{Chang, Chen, and
  Hung}{2011}]{chang2011ordinal}
Chang, K.-Y.; Chen, C.-S.; and Hung, Y.-P.
\newblock 2011.
\newblock Ordinal hyperplanes ranker with cost sensitivities for age
  estimation.
\newblock In {\em CVPR},  585--592.

\bibitem[\protect\citeauthoryear{Chen \bgroup et al\mbox.\egroup
  }{2017}]{chen2017beyond}
Chen, W.; Chen, X.; Zhang, J.; and Huang, K.
\newblock 2017.
\newblock Beyond triplet loss: a deep quadruplet network for person
  re-identification.
\newblock {\em arXiv preprint arXiv:1704.01719}.

\bibitem[\protect\citeauthoryear{Chen, Chen, and Hsu}{2014}]{chen2014cross}
Chen, B.-C.; Chen, C.-S.; and Hsu, W.~H.
\newblock 2014.
\newblock Cross-age reference coding for age-invariant face recognition and
  retrieval.
\newblock In {\em European Conference on Computer Vision},  768--783.
\newblock Springer.

\bibitem[\protect\citeauthoryear{Cootes \bgroup et al\mbox.\egroup
  }{2001}]{cootes2001active}
Cootes, T.~F.; Edwards, G.~J.; Taylor, C.~J.; et~al.
\newblock 2001.
\newblock Active appearance models.
\newblock {\em IEEE TPAMI} 23(6):681--685.

\bibitem[\protect\citeauthoryear{Dalal and Triggs}{2005}]{dalal2005histograms}
Dalal, N., and Triggs, B.
\newblock 2005.
\newblock Histograms of oriented gradients for human detection.
\newblock In {\em CVPR}, volume~1,  886--893.

\bibitem[\protect\citeauthoryear{Geng, Smith-Miles, and
  Zhou}{2008}]{geng2008facial}
Geng, X.; Smith-Miles, K.; and Zhou, Z.-H.
\newblock 2008.
\newblock Facial age estimation by nonlinear aging pattern subspace.
\newblock In {\em Proceedings of the 16th ACM international conference on
  Multimedia},  721--724.
\newblock ACM.

\bibitem[\protect\citeauthoryear{Guo \bgroup et al\mbox.\egroup
  }{2017}]{guo2017zero}
Guo, Y.; Ding, G.; Han, J.; and Gao, Y.
\newblock 2017.
\newblock Zero-shot learning with transferred samples.
\newblock {\em IEEE TIP}.

\bibitem[\protect\citeauthoryear{Guo, Ding, and Han}{2017}]{guo2017robust}
Guo, Y.; Ding, G.; and Han, J.
\newblock 2017.
\newblock Robust quantization for general similarity search.
\newblock {\em IEEE TIP}.

\bibitem[\protect\citeauthoryear{Hadsell, Chopra, and
  LeCun}{2006}]{hadsell2006dimensionality}
Hadsell, R.; Chopra, S.; and LeCun, Y.
\newblock 2006.
\newblock Dimensionality reduction by learning an invariant mapping.
\newblock In {\em CVPR}, volume~2,  1735--1742.

\bibitem[\protect\citeauthoryear{Han, Otto, and Jain}{2013}]{han2013age}
Han, H.; Otto, C.; and Jain, A.~K.
\newblock 2013.
\newblock Age estimation from face images: Human vs. machine performance.
\newblock In {\em Biometrics (ICB), 2013 International Conference on},  1--8.

\bibitem[\protect\citeauthoryear{He \bgroup et al\mbox.\egroup
  }{2005}]{he2005face}
He, X.; Yan, S.; Hu, Y.; Niyogi, P.; and Zhang, H.-J.
\newblock 2005.
\newblock Face recognition using laplacianfaces.
\newblock {\em IEEE TPAMI} 27(3):328--340.

\bibitem[\protect\citeauthoryear{Jain, Klare, and Park}{2012}]{jain2012face}
Jain, A.~K.; Klare, B.; and Park, U.
\newblock 2012.
\newblock Face matching and retrieval in forensics applications.
\newblock {\em IEEE multimedia} 19(1):20.

\bibitem[\protect\citeauthoryear{Kwon and Lobo}{1994}]{kwon1994age}
Kwon, Y.~H., and Lobo, N. D.~V.
\newblock 1994.
\newblock Age classification from facial images.
\newblock In {\em CVPR},  762--767.

\bibitem[\protect\citeauthoryear{Lanitis and Cootes}{2002}]{lanitis2002fg}
Lanitis, A., and Cootes, T.
\newblock 2002.
\newblock Fg-net aging data base.
\newblock {\em Cyprus College}.

\bibitem[\protect\citeauthoryear{Lin, Li, and
  Tang}{2017}]{lin2017discriminative}
Lin, J.; Li, Z.; and Tang, J.
\newblock 2017.
\newblock Discriminative deep hashing for scalable face image retrieval.
\newblock In {\em Proceedings of International Joint Conference on Artificial
  Intelligence}.

\bibitem[\protect\citeauthoryear{Liu and Wechsler}{2002}]{liu2002gabor}
Liu, C., and Wechsler, H.
\newblock 2002.
\newblock Gabor feature based classification using the enhanced fisher linear
  discriminant model for face recognition.
\newblock {\em IEEE TIP} 11(4):467--476.

\bibitem[\protect\citeauthoryear{Luo \bgroup et al\mbox.\egroup
  }{2016}]{luo2016tree}
Luo, L.; Chen, L.; Yang, J.; Qian, J.; and Zhang, B.
\newblock 2016.
\newblock Tree-structured nuclear norm approximation with applications to
  robust face recognition.
\newblock {\em IEEE TIP} 25(12):5757--5767.

\bibitem[\protect\citeauthoryear{Luo \bgroup et al\mbox.\egroup
  }{2017}]{luo2017robust}
Luo, L.; Yang, J.; Qian, J.; Tai, Y.; and Lu, G.-F.
\newblock 2017.
\newblock Robust image regression based on the extended matrix variate power
  exponential distribution of dependent noise.
\newblock {\em IEEE transactions on neural networks and learning systems}
  28(9):2168--2182.

\bibitem[\protect\citeauthoryear{Ou \bgroup et al\mbox.\egroup
  }{2014}]{ou2014robust}
Ou, W.; You, X.; Tao, D.; Zhang, P.; Tang, Y.; and Zhu, Z.
\newblock 2014.
\newblock Robust face recognition via occlusion dictionary learning.
\newblock {\em Pattern Recognition} 47(4):1559--1572.

\bibitem[\protect\citeauthoryear{Ricanek and Tesafaye}{2006}]{ricanek2006morph}
Ricanek, K., and Tesafaye, T.
\newblock 2006.
\newblock Morph: A longitudinal image database of normal adult age-progression.
\newblock In {\em 7th International Conference on Automatic Face and Gesture
  Recognition (FGR06)},  341--345.

\bibitem[\protect\citeauthoryear{Roweis and Saul}{2000}]{roweis2000nonlinear}
Roweis, S.~T., and Saul, L.~K.
\newblock 2000.
\newblock Nonlinear dimensionality reduction by locally linear embedding.
\newblock {\em Science} 290(5500):2323--2326.

\bibitem[\protect\citeauthoryear{Schroff, Kalenichenko, and
  Philbin}{2015}]{schroff2015facenet}
Schroff, F.; Kalenichenko, D.; and Philbin, J.
\newblock 2015.
\newblock Facenet: A unified embedding for face recognition and clustering.
\newblock In {\em CVPR},  815--823.

\bibitem[\protect\citeauthoryear{Shalev-Shwartz, Singer, and
  Ng}{2004}]{Shai2004}
Shalev-Shwartz, S.; Singer, Y.; and Ng, A.~Y.
\newblock 2004.
\newblock Online and batch learning of pseudo-metrics.
\newblock In {\em International Conference on Machine Learning}.

\bibitem[\protect\citeauthoryear{Sun \bgroup et al\mbox.\egroup
  }{2014}]{NIPS2014_5416}
Sun, Y.; Chen, Y.; Wang, X.; and Tang, X.
\newblock 2014.
\newblock Deep learning face representation by joint
  identification-verification.
\newblock In {\em NIPS}.
\newblock  1988--1996.

\bibitem[\protect\citeauthoryear{Suo \bgroup et al\mbox.\egroup
  }{2007}]{suo2007multi}
Suo, J.; Min, F.; Zhu, S.; Shan, S.; and Chen, X.
\newblock 2007.
\newblock A multi-resolution dynamic model for face aging simulation.
\newblock In {\em CVPR},  1--8.

\bibitem[\protect\citeauthoryear{Taigman \bgroup et al\mbox.\egroup
  }{2014}]{taigman2014deepface}
Taigman, Y.; Yang, M.; Ranzato, M.; and Wolf, L.
\newblock 2014.
\newblock Deepface: Closing the gap to human-level performance in face
  verification.
\newblock In {\em CVPR},  1701--1708.

\bibitem[\protect\citeauthoryear{Tang and Wang}{2002}]{tang2002face}
Tang, X., and Wang, X.
\newblock 2002.
\newblock Face photo recognition using sketch.
\newblock In {\em 2002 International Conference on Image Processing}, volume~1,
   I--257.

\bibitem[\protect\citeauthoryear{Wang \bgroup et al\mbox.\egroup
  }{2014}]{wang2014learning}
Wang, J.; Song, Y.; Leung, T.; Rosenberg, C.; Wang, J.; Philbin, J.; Chen, B.;
  and Wu, Y.
\newblock 2014.
\newblock Learning fine-grained image similarity with deep ranking.
\newblock In {\em CVPR},  1386--1393.

\bibitem[\protect\citeauthoryear{Zhao, Han, and
  Shao}{2017}]{zhao2017unconstrained}
Zhao, J.; Han, J.; and Shao, L.
\newblock 2017.
\newblock Unconstrained face recognition using a set-to-set distance measure on
  deep learned features.
\newblock {\em IEEE Transactions on Circuits and Systems for Video Technology}.

\bibitem[\protect\citeauthoryear{Zheng and Shao}{2016}]{zheng2016learning}
Zheng, F., and Shao, L.
\newblock 2016.
\newblock Learning cross-view binary identities for fast person
  re-identification.
\newblock In {\em Proceedings of International Joint Conference on Artificial
  Intelligence},  2399--2406.

\bibitem[\protect\citeauthoryear{Zheng \bgroup et al\mbox.\egroup
  }{2013}]{zheng2013semi}
Zheng, F.; Song, Z.; Shao, L.; Chung, R.; Jia, K.; and Wu, X.
\newblock 2013.
\newblock A semi-supervised approach for dimensionality reduction with
  distributional similarity.
\newblock {\em Neurocomputing} 103:210--221.

\bibitem[\protect\citeauthoryear{Zheng, Tang, and Shao}{2016}]{zheng2016hetero}
Zheng, F.; Tang, Y.; and Shao, L.
\newblock 2016.
\newblock Hetero-manifold regularisation for cross-modal hashing.
\newblock {\em IEEE TPAMI}.

\end{thebibliography}
\bibliographystyle{aaai}
\end{document}